\documentclass{article}
\PassOptionsToPackage{numbers,compress}{natbib}
\usepackage[preprint]{neurips_2026}
\usepackage[utf8]{inputenc}
\usepackage[T1]{fontenc}
\usepackage{amsmath,amssymb,amsfonts}
\usepackage{graphicx,booktabs,array,tabularx,adjustbox}
\usepackage{microtype,xurl,placeins,float}
\usepackage{xcolor}
\usepackage{hyperref}
\usepackage{bookmark}
\hypersetup{hidelinks,pdftitle={StrictPRMBench: A Replayable Evaluation Benchmark for Auditing Fixed-Score PRM Aggregation},pdfauthor={Pengfei Zheng},pdfsubject={Fixed-evidence process reward model aggregation and omission robustness},pdfkeywords={process reward models, aggregation, evaluation, omission robustness, reproducibility}}
\newcolumntype{Y}{>{\raggedright\arraybackslash}X}
\newcolumntype{L}[1]{>{\raggedright\arraybackslash}p{#1}}
\newcommand{\ours}{\textsc{StrictPRMBench}}
\newcommand{\estci}[2]{\begin{tabular}[t]{@{}r@{}}#1\\{}#2\end{tabular}}
\newcommand{\filepath}[1]{\path{#1}}
\title{StrictPRMBench:\\A Replayable Evaluation Benchmark for\\Auditing Fixed-Score PRM Aggregation}
\author{
  Pengfei Zheng\\
  Independent Researcher, Hangzhou, China\\
  \texttt{pengfeizheng21@gmail.com}
}

\date{}
\begin{document}
\maketitle

\begin{abstract}
Process reward models (PRMs) score reasoning steps, but best-of-$N$ selection also requires a rule that aggregates those scores. We study that rule as a separate evaluation object. \ours{} fixes mathematical reasoning pools, scorer evidence, and the source trace of a deletion intervention, so that different reducers face the same recorded selection problem. Across seven scoring checkpoints and two benchmarks, the stored $N=32$ outcomes give higher deletion-induced selection instability for product than geometric mean (GM) in all 14 cells. The equal-cell difference is $+0.0965$; a shared-problem bootstrap, conditional on these checkpoints and frozen pools, gives a 95\% interval of $[0.0846,0.1086]$. A pre-injection-defined sensitivity excludes replacement slots that are either reducer's clean winner and retains 4,459 of 4,622 cell--problem records. The retained difference is $+0.0986$ $[0.0875,0.1098]$. Effects are heterogeneous: the two Qwen cells have positive percentile-bootstrap intervals on retained support but two-sided exact paired $p$-values of 0.0625 and 0.0703. Clean-pool Qwen accuracy does not detect a product advantage, but this is not an equivalence result. We provide versioned outcome records, CPU reanalysis, and an explicit evidence-availability contract. The contribution is a diagnostic evaluation protocol and a reproducible audit of recorded selector behavior, not a new PRM, a universally optimal reducer, or evidence of deployment-wide harm. Raw-score replay and outcome-record reanalysis are distinguished: missing primary edited-score vectors limit independent evaluation of new reducers under the same stress.
\end{abstract}

\section{Introduction}
\label{sec:intro}

\paragraph{The step scorer is not the complete selector.}
A process reward model assigns scores to intermediate reasoning steps. To choose one response from a pool, an inference pipeline must also aggregate the scores of each response and rank the resulting trajectory scores. Product, geometric mean, minimum, and last-step aggregation can produce different rankings from identical step-score vectors. A change in the selected reasoning trace need not change the correctness of its final answer. We therefore distinguish the \emph{step scorer}, the \emph{trajectory reducer}, and the induced \emph{pool-level selector}.

\ours{} asks a narrower question than whether one PRM is generally better than another: when the candidate pool and scored evidence are fixed, how does the reducer affect selection under a verified omission intervention? Our main case study compares product and geometric mean (GM). The algebraic length sensitivity of product is known; it is not our novelty claim. What requires measurement is its selector-level consequence after the edited trace has been rescored and competes with other candidates. That consequence depends on the scorer response, competitor scores, tie-breaking, intervention source, and pool construction.

A useful audit must also distinguish process sensitivity from task failure. Deleting a logically necessary step can damage an explicit derivation while leaving its final answer unchanged. Conversely, deleting a restatement can change the selector without damaging the argument. Thus an arbitrary selection flip is not, by itself, proof of reasoning degradation. We interpret the recorded flips together with the intended edit semantics and separately report what the available correctness and human-evaluation evidence can establish.

\begin{figure}[t]
\centering\includegraphics[width=\linewidth]{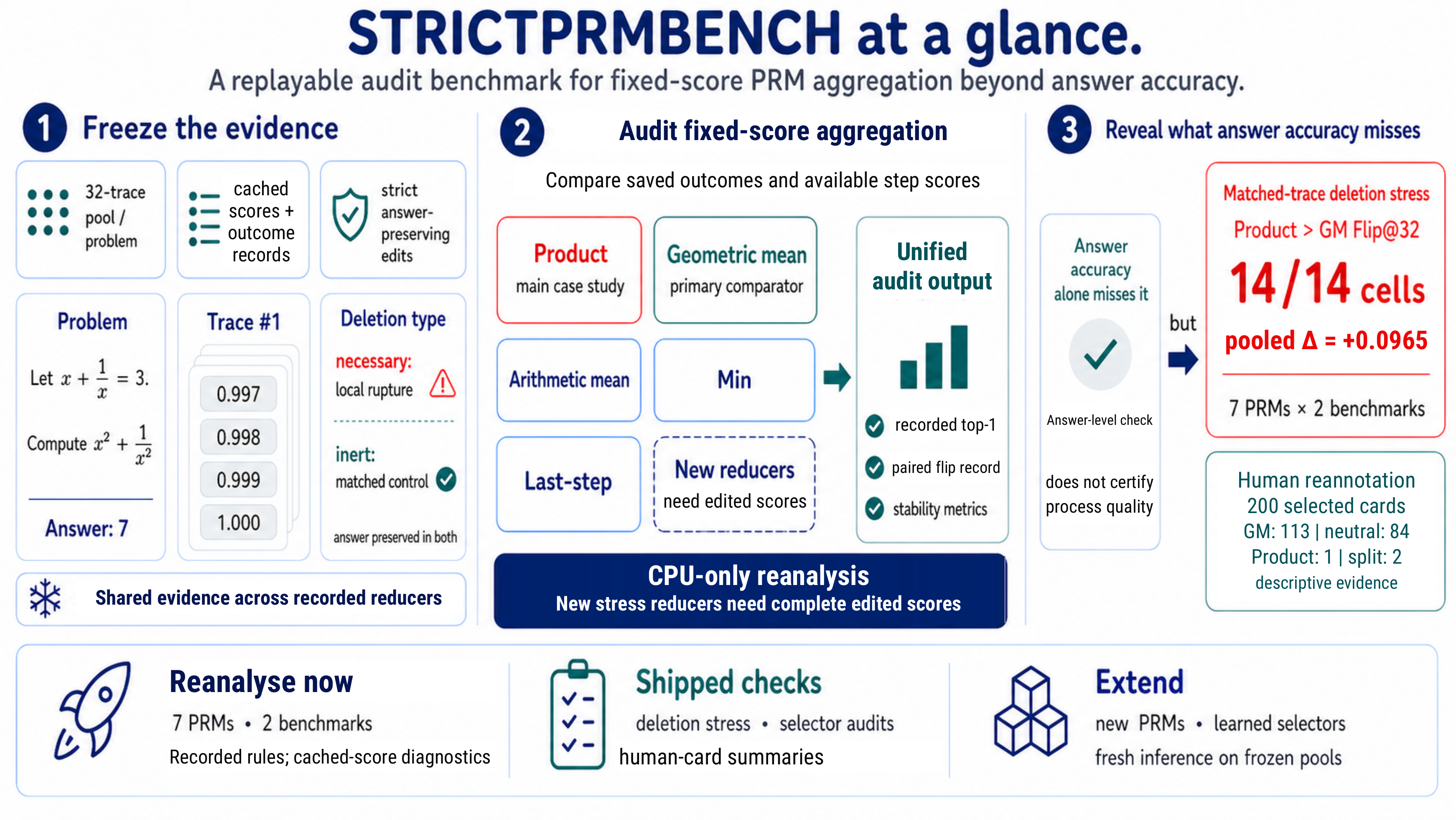}
\caption{\textbf{\ours{} overview and evidence boundary.} The scorer, candidate pool, and intervention are held fixed when comparing reducers. The 14/14 headline counts positive point estimates, not statistically significant cells. Reducer boxes illustrate the interface; full stress replay of a new rule requires the missing primary edited-score vectors. The human counts describe a selected 200-card subset, not deployment prevalence. The example at left is illustrative.}
\label{fig:overview}
\end{figure}

\paragraph{What is established.}
The recorded product-minus-GM Flip@32 difference is positive in all 14 checkpoint--benchmark cells. Eleven cells reject equal flip probabilities under a two-sided exact paired test with Holm correction across the 14-cell family. A replacement-slot sensitivity preserves the positive point-estimate pattern, but the modern Qwen results remain small and sensitive to the inferential procedure. The natural-pool comparison measures clean-answer utility separately from deletion stress. These results justify auditing aggregation as a component; they do not establish that optimizing omission stability improves held-out answer accuracy.

\paragraph{Contributions.}
\begin{enumerate}\setlength\itemsep{3pt}
\item \textbf{An explicit evaluation object.} We separate scorer, reducer, and selector and define a fixed-evidence comparison that attributes differences between recorded selection decisions to the reducer, conditional on the pool and intervention.
\item \textbf{A controlled omission-stress protocol.} We distinguish necessary and inert deletions, fix the attacked source trace across reducers, and identify a comparison-safe retained support that rules out removal of either compared clean winner as the cause of a flip.
\item \textbf{A heterogeneous empirical audit.} We report the full 14-cell product--GM comparison, min/last-step comparisons, a Qwen stability--utility analysis, MATH difficulty strata, and a bounded 72B pilot. Null and inconclusive results remain visible.
\item \textbf{A versioned reanalysis artifact.} The accompanying files pin input records and regenerate the numerical tables on CPU. The release contract explicitly separates verified record-level calculations, historical summaries, and claims for which the necessary underlying evidence is unavailable.
\end{enumerate}

\paragraph{Scope of this public version.}
This is a math-first diagnostic benchmark, not a prevalence study or an end-to-end replication of a deployed Qwen pipeline. Reanalysis of stored outcomes is not independent verification of the PRM inference that produced them. Appendix~\ref{app:status} identifies the claims narrowed or withheld in this version, including unrecovered multi-seed and $N=8$ analyses and an expert-replication independence conflict. These limits are part of the evaluation contract rather than hidden in an unconditional reproducibility claim.

\section{Related Work}
\label{sec:related}
PRMs support stepwise verification, search, and process supervision~\citep{lightman2023lets,wang2024mathshepherd,xie2024montecarlo}. Benchmarks such as ProcessBench and PRMBench study process-error detection and reward-model judgments~\citep{processbench2024,prmbench2025}; tool-use and structural-reasoning evaluations extend this perspective~\citep{toolprmbench2026,hcrs2026}. Our object is the downstream aggregation rule on fixed scorer evidence, not a replacement benchmark for every aspect of PRM quality.

The length bias of unnormalized products and log-sums is established in sequence scoring~\citep{wu2016gnmt,murray2018correcting}. CoLD studies counterfactual length debiasing in PRMs~\citep{cold2026}; other work addresses test-time compute allocation, aggregation, and verifier training~\citep{optimalagg2026,snell2025scaling,prmlessons2025}. These are related questions, not evidence that every recent method was designed as a direct response to product aggregation. In particular, geometric-mean policy optimization concerns the policy-optimization objective~\citep{gmpo2026}, min-form credit assignment is studied in PURE~\citep{pure2025}, and ReasonFlux studies trajectory-aware process reward modeling~\citep{reasonflux2025}. We do not relabel these as validations of our reducer benchmark.

Reward robustness, overoptimization, adversarial training, and transformed-input evaluations motivate stress testing~\citep{rewordbench2025,rewardunderattack2026,brokencompass2026,aprm2025,skalse2022defining,gao2023scaling,coste2024reward,rewardbench2}. Work on theoretical process supervision, automatic process rewards, and alternative verifiers offers complementary approaches~\citep{dostepbystep2025,freeprocessrewards2025,isprmneeded2025,rewardingprogress2025,thinkprm2025}. Evaluating reasoning beyond answer accuracy is also a broader research direction~\citep{reasoneval2024,hard2verify2025,outcomeaccuracy2026}. Our contribution is the particular controlled selector audit, not the general observation that process quality matters.

Qwen documents a product-scored BoN evaluation with eight responses sampled from Qwen2.5-Math-7B-Instruct~\citep{qwen25mathprm2025,qwenprmblog2024}. Our retained Qwen analysis instead uses $N=32$ QwQ-generated pools and the benchmark's matched-trace intervention. It is therefore a documented-component audit at a different operating point, not a replication of the published $N=8$ pipeline. Clear documentation of this distinction follows the motivation of benchmark cards and datasheets~\citep{benchmarkcards2025,constructvalidity2025,gebru2021datasheets}.

\section{The StrictPRMBench Evaluation Protocol}
\label{sec:protocol}

\subsection{Objects, pools, and scoring evidence}
Let $\mathcal P_i=\{z_{ij}\}_{j=1}^{N}$ be a frozen response pool for problem $i$. A scorer checkpoint $m$ assigns a vector $s_{mij}$ to the reasoning steps of each response. A reducer $r$ maps that vector to a scalar, and the selector returns
\begin{equation}
 j_{mir}^{\mathrm{clean}}=\arg\max_j r(s_{mij}),
\end{equation}
using a fixed tie rule. The paired comparison keeps the same pool, scorer checkpoint, source trace, and edited evidence for both reducers. It changes the reduction function, not the generated candidate set.

The benchmark construction uses GSM8K and MATH-500~\citep{cobbe2021training,hendrycks2021math}, QwQ-32B-Preview generated pools, and seven scoring checkpoints. The original release describes 1,819 problem-level pools (1,319 GSM8K and 500 MATH-500) and a 500-problem-per-benchmark working subset. The outcome records used here cover 4,622 checkpoint--problem observations: 2,556 GSM8K observations over 454 distinct questions and 2,066 MATH-500 observations over 393 distinct questions. These are \emph{847 distinct questions}, not 4,622 independent questions. Availability and eligibility differ by checkpoint.

The seven checkpoints are Math-Shepherd, Skywork-7B, Skywork-1.5B, Qwen-7B, RLHFlow-8B, and prefix-scoring adaptations of InternLM2-7B and InternLM2-1.8B. We do not equate an adapted scalar reward model with a natively trained step scorer. Nor do we interpret a product baseline for every checkpoint as its officially recommended deployment configuration. Appendix~\ref{app:models} records these distinctions.

\subsection{Strict edits and two different pairing designs}
The intended necessary-deletion intervention removes one reasoning step whose absence creates a locally checkable gap in the explicit derivation; an inert deletion removes a step without creating such a gap. Answer preservation is checked separately. It concerns the explicit final answer, not an assertion that all reasoning remains valid. A strict edit permits the designated deletion and no unrelated text changes. Appendix~\ref{app:construction} states the validation contract and its semantic limits.

Two uses of ``matched'' must not be conflated. In a \emph{matched-trace reducer comparison}, both reducers see the same injected candidate pool and the same attacked source trace. In the \emph{necessary/inert control comparison}, the construction matches two deletion examples by benchmark, difficulty bucket, deletion position, and deleted-step token length. The latter is generally a \emph{cross-question} match; it does not require the same question, source trace, or total trajectory length. Thus the NI control is not a within-question randomized intervention and cannot, by matching alone, remove all semantic or difficulty confounding.

For necessary/inert matching, difficulty buckets are based on correct-trace-count quintiles in the original candidate pools, rather than the standard MATH Level 1--5 labels. The matching procedure prioritizes agreement in deletion position and deleted-step token length within a bucket, with the documented tolerances and fallback described in Appendix~\ref{app:construction}. The released NI summary contains 200 pairs per benchmark. A further score-matching procedure rematches examples by a deleted-step score proxy; its retained counts vary by checkpoint.

\subsection{Actual replacement rule and comparison-safe support}
\label{sec:safe}
The attacked source trace is selected by the common harness default, so it need not be the product reducer's own clean winner. The actual replacement slot is the largest candidate index other than the harness-default clean top-1. With zero-based indices and $N=32$, it is index 31 unless the harness-default winner is 31, in which case it is 30. The earlier description ``lowest-scored non-top-1'' did not describe this implementation. The reported source results use the last-slot implementation; the prose is corrected here.

Replacing a candidate can itself remove another reducer's clean winner. We therefore define, separately for product versus comparator $c$, the pre-injection support
\begin{equation}
 E_{im}^{(c)}=\mathbf 1\{a_{im}\ne j_{miP}^{\mathrm{clean}},\;
 a_{im}\ne j_{mic}^{\mathrm{clean}}\},
\end{equation}
where $a_{im}$ is the recorded replacement slot. The recorded removal-only selection must also equal each reducer's clean selection. Eligibility does not use a post-injection flip, edited score, correctness label, or desired result. Because the rule uses the compared selectors' clean identities, the support is comparator-specific and is not independent of all selection outcomes.

On retained records, a change to the edited slot cannot be explained solely by deleting the prior winner. This statement applies \emph{only to retained support}. It neither repairs the historical replacement implementation nor establishes append-only, lowest-score-replacement, or universally safe replacement robustness. Those alternative protocols require the missing edited-score inputs and a separate evaluation.

\subsection{Metrics and estimands}
For a recorded intervention, define $F_{mir}=\mathbf1\{j_{mir}^{\mathrm{inj}}\ne j_{mir}^{\mathrm{clean}}\}$. Flip@32 is its mean on the specified support. The paired difference is
\begin{equation}
 d_{mi}^{(c)}=F_{miP}-F_{mic},\qquad
 \widehat\Delta^{(c)}=\frac1{14}\sum_{m,b}\frac1{n_{mb}^{(c)}}\sum_{i\in I_{mb}^{(c)}}d_{mi}^{(c)}.
\label{eq:macro}
\end{equation}
Every checkpoint--benchmark cell receives equal weight. This is not the row-weighted average over all observations. Positive values indicate more flips under product; they do not imply that every flip is harmful or that the comparator is universally optimal.

APR is the fraction of clean/edited trace pairs for which the edited trace receives a higher reducer score. The NI statistic is $\Delta_{\mathrm{NI}}=\mathrm{APR}_{\mathrm{necessary}}-\mathrm{APR}_{\mathrm{inert}}$ on a stated matching scheme. The published raw and score-matched NI point estimates (14 cells each) are recomputed exactly from the public per-pair score records. Their historical intervals resampled pair rows even though questions recur across pairs, so we replace them with question-cluster bootstrap intervals (Appendix~\ref{app:ni}).

Wrong@32 concerns a clean-correct selection becoming answer-incorrect after injection. It is not the overall error rate of the natural pools. Historical summaries report zero across 84 selector-by-cell settings, but the present record archive lacks the primary edited-answer correctness needed to recompute all of those indicators independently. We consequently treat that zero as a historical scope observation, not new correctness validation or proof that the stress is harmless.

Natural-pool Gain@32 is selected-answer correctness minus the paired random-candidate baseline on the \emph{unedited} pools. For two reducers using the same baseline, $\Delta$Gain equals their accuracy difference. Its support is not restricted to stress-eligible or comparison-safe records.

\subsection{Inference and reproducibility levels}
All new calculations use 10,000 bootstrap replicates and PCG64 seed 42. Within a single cell, each problem appears once and we bootstrap the paired difference $d$. Across cells, we resample problem IDs jointly within each benchmark, carrying all available checkpoint observations for a sampled question together, and recompute the equal-cell mean. For the NI controls, where one question can supply several deletion instances, we resample question IDs within each cell and carry all of a question's necessary and inert instances together. The checkpoint suite, candidate generation, matching, and eligibility masks remain fixed. These CIs quantify conditional question-level sampling variability, not variation over new checkpoints, training runs, generators, or new edit constructions.

For binary paired outcomes we also report a two-sided exact discordance test, equivalent to the exact McNemar/binomial test. Holm adjustment is applied across the 14 cells of each declared comparator/support family. The product--GM family is primary; min and last-step comparisons are secondary. The reported percentile intervals are pointwise, not simultaneous confidence intervals. Small discordant counts can make approximate bootstrap intervals disagree with exact tests; we report rather than conceal that disagreement.

There are three evidence levels. \emph{Score-vector replay} applies a new reducer where the required score vectors exist. \emph{Outcome-record reanalysis}, which is fully packaged for the main results here, recomputes statistics for recorded reducers without new inference. \emph{Summary-only evidence} cannot establish independent recovery of the underlying annotations or missing model outputs. Appendix~\ref{app:artifact} makes this contract executable.

\section{Results}
\label{sec:results}

\subsection{Product--GM differences persist on fixed evidence}
\begin{figure}[t]
\centering\includegraphics[width=\linewidth]{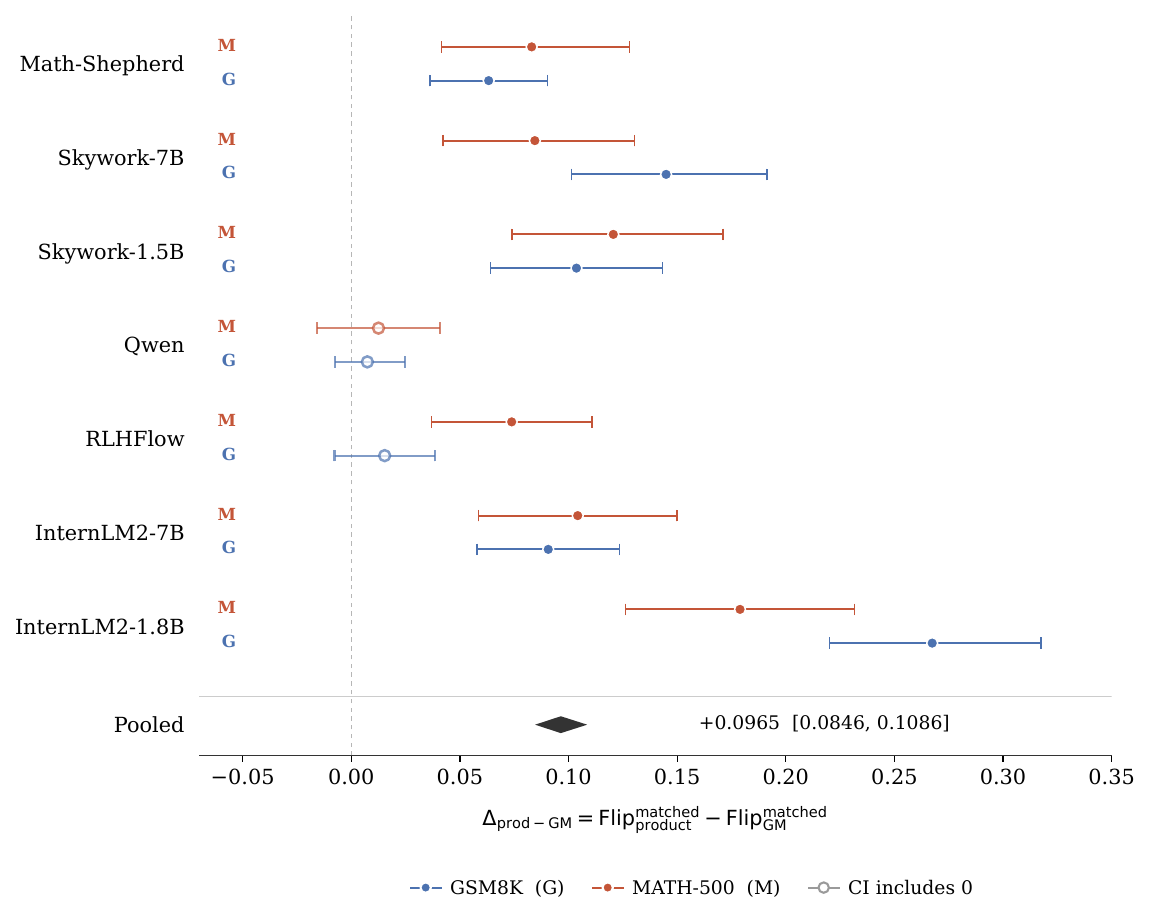}
\caption{\textbf{Product-minus-GM Flip@32 on full recorded support.} Every point, interval, and appendix row is generated from the same paired outcome records. Cell intervals use a paired problem bootstrap; the pooled row uses the shared-problem bootstrap conditional on the fixed 14-cell suite. The plotted positive direction in 14/14 cells is not a claim of significance in all cells. Table~\ref{tab:cellsfull} reports exact paired and Holm-adjusted tests.}
\label{fig:forest}
\end{figure}

Figure~\ref{fig:forest} shows all 14 full-recorded-support cells. Product has the higher flip rate in every cell, with differences ranging from $+0.0074$ (Qwen/GSM8K) to $+0.2674$ (InternLM2-1.8B/GSM8K). The equal-cell mean is $+0.0964616$, or 9.646 percentage points. Eleven of the 14 cells reject under the exact paired tests after Holm adjustment. The pattern is not uniformly large: the strongest adapted-model cell should not stand in for the modern Qwen cells.

\begin{table}[t]
\centering\small\begin{adjustbox}{max width=\linewidth}% Auto-generated; do not edit numbers by hand.
\begin{tabular}{llrrr}
\toprule
Comparator & Support & Cell--problem rows & $\Delta$ [95\% CI] & Holm rejects \\
\midrule
GM & Full recorded & 4,622 & \estci{$+0.0965$}{$[+0.0846, +0.1086]$} & 11/14 \\
GM & Comparison-safe & 4,459 & \estci{$+0.0986$}{$[+0.0875, +0.1098]$} & 11/14 \\
Min-step & Full recorded & 4,622 & \estci{$+0.1011$}{$[+0.0894, +0.1132]$} & 14/14 \\
Min-step & Comparison-safe & 4,502 & \estci{$+0.0897$}{$[+0.0785, +0.1008]$} & 9/14 \\
Last-step & Full recorded & 4,622 & \estci{$+0.1270$}{$[+0.1135, +0.1412]$} & 14/14 \\
Last-step & Comparison-safe & 4,470 & \estci{$+0.1228$}{$[+0.1096, +0.1363]$} & 14/14 \\
\bottomrule
\end{tabular}
\end{adjustbox}
\caption{\textbf{Equal-cell selection-instability differences.} All rows compare product with the named comparator. Intervals are the shared-problem, fixed-checkpoint bootstrap. ``Holm rejects'' counts two-sided exact tests rejected at 0.05 within that row's 14-cell family. Different comparison-safe rows have different supports, so their magnitudes are not a ranking on one common population.}
\label{tab:pooled}
\end{table}

The comparison-safe product--GM analysis retains 4,459/4,622 records (96.47\%). Its macro difference is $+0.0986373$ with shared-problem CI $[0.0875,0.1098]$. The positive point-estimate pattern survives in all 14 cells. All retained recorded flips select the actual edited slot, and removal-only leaves both compared selectors unchanged. This excludes the clean-winner-removal explanation on this subset; it does not establish robustness to alternative injection protocols. Although all 14 safe percentile intervals are positive, only 11/14 exact tests remain significant after Holm adjustment.

Min-step and last-step provide additional recorded-selector comparisons (Table~\ref{tab:pooled}). They do not establish performance of arbitrary newly defined reducers. The named Qwen/Skywork six-cell subset has mean $+0.0789$ on full recorded support. We identify its three checkpoints explicitly rather than calling it an all-native-PRM estimate: it omits Math-Shepherd and RLHFlow.

\subsection{Qwen: modest stress effects, unresolved utility trade-off}
\label{sec:qwen}
Qwen is useful because product is a documented aggregation choice, not because every part of our pipeline matches its published evaluation. Table~\ref{tab:qwenstress} reports the comparison-safe $N=32$ stress outcome. GSM8K has five product-only flips and no GM-only flips; MATH-500 has seven product-only and one GM-only flip. The positive percentile intervals coexist with exact two-sided $p=0.0625$ and $p=0.0703$. We therefore describe these as small positive estimates with method-sensitive uncertainty, not unqualified statistically significant superiority.

\begin{table}[htbp]
\centering\small\begin{adjustbox}{max width=\linewidth}% Auto-generated; do not edit numbers by hand.
\begin{tabular}{lrrrrr}
\toprule
Benchmark & $n$ & Flips P/GM & Discordant P/GM & $\Delta$ [95\% CI] & Exact $p$ \\
\midrule
GSM8K & 397 & 12/7 & 5/0 & \estci{$+0.0126$}{$[+0.0025, +0.0252]$} & 0.0625 \\
MATH-500 & 302 & 17/11 & 7/1 & \estci{$+0.0199$}{$[+0.0033, +0.0397]$} & 0.0703 \\
\bottomrule
\end{tabular}
\end{adjustbox}
\caption{\textbf{Qwen comparison-safe stress at $N=32$.} P denotes product. ``Flips'' gives each rule's total; ``discordant'' gives product-only and GM-only flip counts. Exact tests, rather than the positivity of a percentile CI alone, expose the small effective discordance sample.}
\label{tab:qwenstress}
\end{table}

\begin{table}[htbp]
\centering\small\begin{adjustbox}{max width=\linewidth}% Auto-generated; do not edit numbers by hand.
\begin{tabular}{lrrrrr}
\toprule
Benchmark & Acc. P & Acc. GM & $\Delta$Gain [95\% CI] & Correct-only P/GM & Exact $p$ \\
\midrule
GSM8K & 0.894 & 0.894 & \estci{$+0.0000$}{$[-0.0100, +0.0100]$} & 3/3 & 1.000 \\
MATH-500 & 0.744 & 0.738 & \estci{$+0.0060$}{$[-0.0080, +0.0200]$} & 8/5 & 0.581 \\
\bottomrule
\end{tabular}
\end{adjustbox}
\caption{\textbf{Qwen natural-pool accuracy and paired utility.} Each row uses all 500 frozen unedited pools, independently of stress filtering. ``Correct-only'' gives the discordant correctness counts; the exact test is paired. A CI containing zero does not establish equivalence or non-inferiority.}
\label{tab:qwennatural}
\end{table}

On GSM8K, both reducers answer 447/500 questions correctly; on MATH-500, product answers 372/500 and GM 369/500. Product and GM choose different clean traces on 121/500 and 216/500 questions, respectively. Of those disagreements, both selected answers are correct on 107 and 149 questions. This directly demonstrates that answer-level correctness can conceal selector disagreement. It does not establish that those natural-pool disagreements are reasoning-quality failures.

An operator who explicitly values omission stability may prefer GM after examining these estimates. Such a choice depends on a utility/risk preference, not a demonstrated accuracy dominance relation. The MATH-500 utility interval still permits a product advantage of two percentage points. No equivalence margin was tested, no held-out predictive-validity experiment is established, and no $N=8$ deployment recommendation follows. The earlier $N=8$/20-seed summary is withheld from empirical support because its per-problem and seed-level archive is unavailable in this version.

\subsection{Difficulty stratification with aligned denominators}
\label{sec:levels}
The level map covers all 500 MATH-500 questions. This analysis conditions on the recorded flags indicating that \emph{both} product and GM select a correct clean answer. It uses 1,926 checkpoint--problem rows over 376 questions, rather than the 2,066 MATH rows of the full recorded audit. This correctness restriction is distinct from comparison-safe replacement filtering. Table~\ref{tab:levels} computes both marginal flip rates and their difference on exactly this same support, avoiding a subtraction of rates with different denominators.

\begin{table}[htbp]
\centering\small\begin{adjustbox}{max width=\linewidth}% Auto-generated; do not edit numbers by hand.
\begin{tabular}{rrrrrr}
\toprule
Level & Problems/rows & Has wrong candidate & Product & GM & $\Delta$ [95\% CI] \\
\midrule
1 & 39/227 & 36/227 & 0.2240 & 0.1175 & \estci{$+0.1066$}{$[+0.0543, +0.1580]$} \\
2 & 77/435 & 84/435 & 0.1980 & 0.0834 & \estci{$+0.1146$}{$[+0.0757, +0.1565]$} \\
3 & 82/418 & 175/418 & 0.1934 & 0.1024 & \estci{$+0.0910$}{$[+0.0444, +0.1414]$} \\
4 & 93/469 & 179/469 & 0.1676 & 0.0991 & \estci{$+0.0685$}{$[+0.0338, +0.1039]$} \\
5 & 85/377 & 239/377 & 0.1847 & 0.0719 & \estci{$+0.1128$}{$[+0.0680, +0.1595]$} \\
\bottomrule
\end{tabular}
\end{adjustbox}
\caption{\textbf{MATH-500 levels, common clean-correct support.} Marginals are equal-checkpoint means over the same records; their difference equals the paired point estimate before rounding. CIs jointly resample problem IDs within level. ``Has wrong candidate'' counts checkpoint--problem records whose pool contains at least one incorrect candidate; it is not a unique-question prevalence estimate.}
\label{tab:levels}
\end{table}

All five level estimates are positive and their pointwise intervals exclude zero. There is no monotonic difficulty trend, and we do not make a trend-test claim. Incorrect-candidate record counts range from 36/227 at Level 1 to 239/377 at Level 5. Thus these are not universally correct candidate pools. However, absence of edited-answer correctness in the record archive prevents an independent recomputation of level-specific Wrong@32; a zero inherited from a historical aggregate is not a fresh stratum-level verification.

\subsection{Necessary/inert controls support an association, not causal identification}
The raw and score-matched NI point estimates, recomputed from the public per-pair records, are negative in all 14 cells (Table~\ref{tab:ni}). Question-cluster 95\% intervals lie below zero in all 14 raw cells and, under the historical score-matching assignment, in 13 of 14 score-matched cells; the exception is InternLM2-1.8B on MATH-500, $[-0.133,+0.025]$. Thus necessary deletions less often increase the product trajectory score than the matched inert deletions. It does not mean the scorer ignores necessity entirely; nor does it establish that aggregation causes the entire observed NI difference.

The cross-question design does not guarantee equal underlying reasoning difficulty or equal total trajectory length. Moreover, questions recur across pairs (the 400 GSM8K deletion instances in 200 pairs come from 164 questions, the 400 MATH-500 instances from 152), whereas the historical NI script resampled paired-difference rows; the question-cluster intervals are 1.3--1.9 times as wide. The intervals hold fixed which instances enter each arm. For the score-matched estimate this includes the caliper assignment, which is not unique: many assignments attain exactly the optimal cost. Over all optima of the stated exact-arithmetic objective, the matched estimate is negative in 13 cells but ranges from $-0.138$ to $+0.028$ for InternLM2-1.8B on MATH-500. Endpoint-assignment intervals are conditional sensitivities, not worst-case interval bounds (Appendix~\ref{app:ni}). The raw estimates do not depend on this assignment. The five-seed sign pattern and ranges remain withdrawn because the public multi-seed records do not identify seed membership.

\subsection{Human evidence supports a limited interpretation}
\label{sec:human}
The 200-card three-choice reannotation summary reports 113 GM-majority, 84 no-preference-majority, one product-majority, and two split outcomes. This is the complete distribution on a subset selected from deletion-stress disagreements, not a population-level preference estimate. It is foregrounded as a descriptive supporting endpoint; it was a later reannotation, not a preregistered primary endpoint. We do not interpret it as a length-independent preference because the compared traces differ in length by construction.

The natural-pool reannotation is a different study: 374/420 cards have a no-preference majority. The stored comparator is preferred on 28 cards, product on 15, and three are split. The resolved-card comparison is not significant at 0.05 under a two-sided exact test. These records argue against equating targeted stress behavior with common spontaneous human-perceived defects. Human tables are transcribed from hash-verified published summaries; this revision has not independently established the identity or independence of every annotator workbook.

The expert mini-replication is \emph{not} used as independent corroboration. Its published input audit records three byte-identical completed workbooks, so the available archive supports at most one distinct return. The former three-independent-expert claims and associated rates are withdrawn. Qwen Likert multi-rater and difference estimates that lack the required completed-workbook set are likewise not used. Appendix~\ref{app:human} preserves the distinction between the usable descriptive summaries and unsupported independence claims.

\subsection{A bounded 72B pilot}
The available Qwen-72B record file contains 143 eligible GSM8K questions from a 200-question sample, not a 600-question two-benchmark experiment. Product flips on 13/143 and GM on 8/143; the difference is $+0.03497$ with paired percentile CI $[0,0.06993]$. There are six product-only and one GM-only flips, giving exact two-sided $p=0.125$. This is an exploratory, inconclusive pilot under the recorded replacement protocol. It does not establish that scaling fails to improve robustness, and no 600-question or cross-benchmark 72B conclusion is retained.

\section{Discussion, Limitations, and Conclusion}
\label{sec:discussion}
\paragraph{What the benchmark adds.}
A step scorer and a selector are different evaluation objects. Holding the pool and scored evidence fixed makes their distinction operational: the stored outcomes show substantial aggregation-dependent selection differences for several checkpoints, while modern Qwen effects are much smaller. This provides a reproducible component-level diagnostic, not a newly discovered length-bias theorem or a generally superior reasoning algorithm.

\paragraph{Why accuracy alone is insufficient, and what that does not prove.}
The same correctness outcome can accompany different selected traces. This motivates process-sensitive evaluation, but it does not by itself prove that the changed trace is worse. Verified edit semantics and human judgments address different pieces of that question; both have selection and provenance limits here. We do not interpret Flip as an unconditional process-quality score, the historical Wrong=0 report as a correctness proof, or a nonsignificant natural accuracy comparison as zero utility cost.

\paragraph{External validity.}
The results condition on two mathematical benchmarks, one candidate generator, a fixed set of checkpoints, and a particular omission construction. Adapted scalar reward models and native PRMs are not interchangeable. Some checkpoints have very small effects. There is no validated outside-math extension, learned-selector evaluation, passive-deployment prevalence estimate, or held-out relation showing that lower deletion instability predicts higher BoN accuracy. No claim is made about a deployed RL policy learning to shorten reasoning.

\paragraph{Artifact limitations and corrections.}
The released outcome records permit exact recounting and statistical reanalysis of recorded selectors. Missing primary edited-score vectors prevent a complete independent stress replay for a newly proposed reducer. Missing annotation and seed-level records prevent stronger claims about expert replication, some Likert analyses, and seed robustness. This public version makes those boundaries explicit, corrects the replacement description and figure/table values, and provides a versioned claim ledger rather than treating successful compilation or checksum agreement as scientific verification.

\paragraph{Responsible use.}
Use the benchmark to inspect reducer behavior under a declared intervention and support, alongside task utility and uncertainty. Do not select a reducer solely because one approximate CI excludes zero. New score-based or trajectory-aware selectors require suitable inputs and, where necessary, fresh inference. A credible cross-domain port must redefine the process unit, semantic omission test, preservation oracle, and environment replay contract; a model scored outside its domain is not automatically a validated extension.

\paragraph{Conclusion.}
\ours{} treats aggregation as an auditable component of PRM-based selection. Its retained evidence supports a conditional finding: on fixed mathematical pools, reducers can differ in omission-induced selection stability even when answer-level statistics do not resolve their utility differences. A transparent separation of protocol, stored outcomes, inference, and evidence availability is necessary for that finding to be reproducible and appropriately scoped.

\paragraph{Data and code availability.}
The original snapshot is in \url{https://github.com/staged-benchmark-2026/StrictPRMBench} at commit \texttt{2ad65ed}; the corresponding dataset is in \url{https://huggingface.co/datasets/staged-benchmark-2026/StrictPRMBench}. The \texttt{anc/} directory accompanying this version packages pinned selector records, descriptive summaries, provenance, and the CPU scripts used for the revised tables. It is a versioned reanalysis layer, not a claim that new files were present in the original snapshot. Appendix~\ref{app:artifact} documents exact contents and boundaries.

\clearpage
\appendix
\section{Aggregation Algebra and Its Limits}
\label{app:algebra}
This appendix states the mechanical effect separately from fresh rescoring. It is not an originality claim. Let a trace have $T>1$ positive scores $s_t\in(0,1]$ and write $a_t=\log s_t$. For implementation, scores are clipped below at $\varepsilon=10^{-30}$ before taking logs; the algebra below then concerns the clipped values.

\paragraph{Product.}
If deletion removes step $k$ without changing the remaining scores,
\begin{equation}
 \log P(s_{-k})-\log P(s)=-a_k.
\end{equation}
This is positive for $s_k<1$ regardless of whether the deleted text was necessary. The gain is a numerical property of multiplying fewer factors below one. It does not establish how a PRM will rescore the shortened context or whether the edited trace will beat its competitors.

\paragraph{Geometric mean.}
Writing $\bar a=T^{-1}\sum_t a_t$, the corresponding change in log GM is
\begin{equation}
 \log G(s_{-k})-\log G(s)=\frac{\bar a-a_k}{T-1}.
\end{equation}
It is positive exactly when the deleted score is below the original geometric mean, zero at equality, and negative above it. Thus length normalization does not make GM immune to deletions.

\paragraph{Minimum and last-step.}
Removing a step increases the minimum only when it is the \emph{unique} minimum (for $T>1$). Removing one of several tied minima leaves the minimum unchanged. Last-step aggregation is invariant to non-final deletion only in the mechanical fixed-score calculation; fresh rescoring can change the last step's score. These are not unconditional robustness guarantees.

\paragraph{Length-normalized family.}
For $\alpha\in\mathbb R$, let $L_\alpha(s)=T^{-\alpha}\sum_t a_t$. Product and GM rankings correspond to $\alpha=0$ and $\alpha=1$. A mechanical deletion gives
\begin{equation}
 L_\alpha(s_{-k})-L_\alpha(s)
 =\big((T-1)^{-\alpha}-T^{-\alpha}\big)\sum_t a_t-(T-1)^{-\alpha}a_k.
\end{equation}
If \emph{every candidate in a comparison} has the same length, product and GM rank those candidates identically. Preserving the length of one clean/edited pair does not imply identical rankings across an otherwise variable-length pool. A placeholder control must state which of these two conditions it enforces.

\paragraph{Fresh rescoring.}
If the remaining scores after deletion are $a'_t$, the product log-score change is
\begin{equation}
 -a_k+\sum_{t\ne k}(a'_t-a_t).
\end{equation}
The second term can offset or exceed the mechanical bonus. Pool-level flips depend additionally on competing traces. Consequently neither the sign nor the size of every selector comparison is an algebraic inevitability. This version does not use unsupported mechanistic-decomposition coefficients to quantify the rescoring term.

\begin{figure}[htbp]
\centering
\fbox{%
\begin{minipage}{0.86\linewidth}
\centering\small
\begin{tabular}{@{}lccc@{}}
\toprule
Trace & Step scores & Product & GM \\
\midrule
Clean & $(0.9, 0.8, 0.7, 0.9)$ & $0.4536$ & $0.8207$ \\
Delete the $0.7$ step & $(0.9, 0.8, 0.9)$ & $0.6480$ & $0.8653$ \\
\bottomrule
\end{tabular}
\vspace{0.35em}

Product gain: $42.9\%$; GM gain: $5.4\%$ (relative changes).
\end{minipage}%
}
\caption{\textbf{The deletion paradox in a simplified example.}
Each number is a fixed step score ($1 =$ perfect). Deleting the $0.7$ term increases both scores, with a larger relative increase for product. This is a mechanical illustration, not a measured PRM result: actual edited traces are freshly rescored, and a score increase alone does not guarantee a pool-level selection flip.}
\label{fig:concept}
\end{figure}

\begin{figure}[htbp]
\centering
\fbox{%
\begin{minipage}{0.92\linewidth}
\small
\textbf{Decomposition readout.}
For a deleted step $k$, product changes through a positive mechanical term and a PRM rescoring term:
\[
\Delta \log S_{\mathrm{prod}} = -\log s_k + \Delta_{\mathrm{ctx}} .
\]
The first term is positive whenever $s_k<1$; $\Delta_{\mathrm{ctx}}$ records whether rescoring of the remaining context compensates for that deletion bonus. The terms require clean and edited step scores. This algebraic panel does not estimate checkpoint-specific compensation or determine whether the edited trace beats its competitors.
\end{minipage}%
}
\caption{\textbf{Mechanical bonus versus rescoring.}
$\Delta_{\mathrm{ctx}}$ is the summed log-score change on the remaining steps; a positive total means the mechanical bonus exceeds the rescoring penalty. This is a decomposition, not a validation of unavailable scatter data or a claim that one checkpoint compensates more strongly than another.}
\label{fig:mech_vs_ctx}
\end{figure}

\FloatBarrier

\section{Construction Contract and Denominator Ledger}
\label{app:construction}
\subsection{Edit semantics and validation}
The construction separates four checks. A \emph{syntactic} check confirms that the designated single span, and no other characters, was deleted. An \emph{answer} check compares the final extracted answer before and after editing under the stated normalization. A \emph{semantic} check decides whether the omission creates a locally verifiable gap. A \emph{scoring} check applies the checkpoint's input formatting and step segmentation consistently to the clean and edited texts. Passing one check does not imply passing the others.

Necessary omissions include a missing calculation needed by a later line, an absent definition on which a later step depends, or a removed logical implication. Inert omissions include redundant restatements or commentary whose removal leaves the explicit argument locally supported. These are labeling guidelines rather than an automatic theorem prover. Ambiguous necessity should be marked and excluded by a documented rule, not silently reassigned to create a stronger effect. The benchmark's semantic conclusions are conditional on these labels; we do not claim that the present outcome-record script independently relabels the original text.

Answer preservation is likewise narrower than semantic preservation. A deleted proof step may leave the final number intact while invalidating the derivation. Conversely, a formatting-only change can perturb a scorer without creating a semantic error. For this reason, a reliable extension should store the original text, exact edited text, a character diff, answer-extraction output, semantic label and rationale, and both score vectors. Counts from a cached selection table cannot replace these records.

\subsection{Necessary/inert matching}
The frozen construction notes specify the same benchmark and difficulty bucket, a deletion-position tolerance of $\pm1$, and a deleted-step token-length tolerance of $\pm30\%$, prioritizing closer matches and using a nearest same-bucket fallback. Difficulty is based on the quintile of correct-trace count, not MATH's annotated difficulty levels. No same-question identity constraint appears in that rule.

The 200 necessary/inert matches per benchmark should therefore be described as cross-question matched controls. Total trace length is not the matching variable. A separate score-matching step uses a caliper of 0.10 on the mechanical bonus proxy $-\log s_k$, with a one-to-one assignment among retained examples. Conditioning on this score can alter the sample and does not create randomization. Inference for the NI statistic therefore conditions on these selections and must respect the dependency structure described next.

A source question or trace can connect more than one cross-question pair, and in the released pairs questions often do. The public per-pair records keep the question ID and candidate (trace) index of both members of every pair. The 400 GSM8K deletion instances (200 pairs) come from 164 questions and 383 traces; 81 questions supply more than one instance (up to 22), and 36 appear in both roles. The 400 MATH-500 deletion instances come from 152 questions and 393 traces; 92 questions supply more than one instance (up to 17). The same 200 pairs per benchmark are used for all seven checkpoints. Resampling rows independently is not equivalent to resampling these connected question-level units, so Appendix~\ref{app:ni} resamples questions. We do not infer independence from the fact that two examples were called a ``pair.''

\subsection{Supports used in this version}
\begin{table}[htbp]
\centering\small
\begin{tabularx}{\linewidth}{L{3.5cm}rY}
\toprule
Analysis & Records & Meaning of denominator\\
\midrule
Full product--GM & 4,622 & 14 cells over 847 distinct benchmark--question IDs; all available paired selector records, not an assertion that every clean winner is correct.\\
Comparison-safe GM & 4,459 & Same source records, excluding replacement of either compared clean winner; 845 distinct questions.\\
Comparison-safe min & 4,502 & Comparator-specific retained support; not the same sample as safe GM.\\
Comparison-safe last & 4,470 & Comparator-specific retained support.\\
Qwen natural utility & 1,000 & 500 unedited pools per benchmark, independent of stress eligibility.\\
MATH level analysis & 1,926 & 376 questions with both compared clean winners recorded as correct; not a comparison-safe filter.\\
72B pilot & 143 & Eligible GSM8K records from a documented 200-question sample.\\
NI controls & 400 & 200 cross-question matched pairs per benchmark; score rematching changes each cell's count.\\
Human reannotation & 200 & Selected deletion-stress disagreement cards; not a random natural-pool sample.\\
Natural human audit & 420 & Cards from a separate natural-pool reannotation summary.\\
\bottomrule
\end{tabularx}
\caption{\textbf{Denominators are part of the estimand.} The same word ``pool'' must not collapse unique questions, checkpoint--question rows, paired edits, and human cards into one sample size.}
\label{tab:supports}
\end{table}

For the MATH level analysis, the stored clean-correct flags define its common support. The original source summary computed the two marginal flip rates on rule-specific clean-correct supports while computing the difference on their intersection. That made some displayed rates incompatible with the paired difference. Here all three entries are recomputed on the intersection. The previous paired point-estimate pattern persists, but several marginal values are corrected.

\paragraph{Wrong@32 provenance.}
The archived difficulty records explicitly describe their Wrong fields as implications of exact-zero historical aggregate cells, not observations reconstructed from edited-answer correctness. The current scripts do not convert this derived field into new evidence. A future independently verified Wrong table requires both the selected candidate identity and its answer correctness, including the edited candidate, together with the clean-correct eligibility definition.

\section{Statistical Analysis Specification}
\label{app:statistics}
\subsection{Per-cell paired inference}
For a fixed checkpoint and benchmark, let $n_{10}$ count questions that flip under product but not under the comparator, and $n_{01}$ the reverse. The effect estimate is $(n_{10}-n_{01})/n$. For the exact test, condition on $n_{10}+n_{01}$ and test a binomial probability of one half, two-sided. With no discordant observations we report $p=1$. The magnitude denominator $n$ and the test's discordance denominator are different and both are reported where useful.

The percentile bootstrap resamples the paired vector $d_i\in\{-1,0,1\}$ with replacement. It does not independently resample product and comparator outcomes. All interval endpoints use the 2.5th and 97.5th percentiles of 10,000 resamples. They are approximate, conditional, pointwise intervals. An empirical sample with few discordances, especially none in one direction, can give a percentile interval excluding zero while a discrete two-sided exact test does not reject. We do not treat one method's label as overriding the other.

\subsection{Shared-problem macro bootstrap}
For each benchmark, create a question-by-checkpoint matrix with the paired difference in observed cells and a parallel presence indicator. Draw multinomial bootstrap question weights with total equal to the number of unique observed questions in that benchmark. The \emph{same weight} is applied to every checkpoint observation for that question. Divide weighted differences by weighted presence counts within each cell, then average all 14 cell means equally. Questions are sampled separately within GSM8K and MATH-500. The implementation fails rather than silently filling an empty cell.

This procedure preserves recorded cross-checkpoint dependence on shared questions. It fixes the seven checkpoints and the two benchmark strata. It does not support inference to a hypothetical population of all PRMs. It also conditions on the already constructed edit set and support; there is no bootstrap regeneration of candidate pools, semantic labels, or score matching. Missing checkpoint--question records are not imputed as zero effects.

An earlier archived summary reported a cell-then-problem hierarchical interval $[0.0630,0.1349]$ around the same macro point $0.0964616$. That procedure resampled cells and problems within cells rather than jointly carrying shared question observations. We retain its provenance in the change record, but use the explicitly fixed-suite shared-problem interval $[0.0846,0.1086]$ in this version. The interval change is a change of inferential target and resampling implementation, \emph{not} an improved model result or new experimental replication.

\subsection{Multiple comparisons and conditional supports}
Holm adjustment is applied separately to each 14-cell comparator/support family. The main family is full-support product versus GM. The comparison-safe family addresses a prespecified form of removal confounding conditional on the recorded clean selections; it is a sensitivity analysis designed during verification, not an independently preregistered experiment. Secondary min and last-step families are labeled as such. We do not claim simultaneous coverage from a collection of pointwise CIs or count 14 cells as 14 independent training runs.

The comparison-safe construction may change which questions are retained for different checkpoints and comparators. Its effect estimates pertain to that retained subset. Reporting retention counts does not prove that excluded questions would have behaved similarly. Appendix~\ref{app:cells} gives both full and retained results, including exact-test non-rejections.

\subsection{Natural utility}
Paired correctness differences are formed on all 500 unedited Qwen pools per benchmark. The random baseline in Gain cancels exactly when computing product minus GM on the same question. The exact correctness test uses product-only-correct and GM-only-correct counts, not stress flip counts. A confidence interval containing zero does not show that switching reducers has no accuracy cost. No non-inferiority or equivalence claim is made without a declared margin and adequate precision.

\begin{figure}[htbp]
\centering
\fbox{%
\begin{minipage}{0.92\linewidth}
\centering\small
\begin{tabular}{@{}p{0.23\linewidth}p{0.66\linewidth}@{}}
\toprule
Quantity & Pattern in the verified stability--utility evidence \\
\midrule
Stability & Product has higher recorded Flip@32 in 14/14 cells; not all differences are statistically significant. \\
Utility & Qwen natural-accuracy intervals include zero; neither a product advantage nor utility equivalence is established. \\
Transition & A product $\to$ GM choice requires a robustness--utility policy, not an assertion of zero accuracy cost. \\
\bottomrule
\end{tabular}
\end{minipage}%
}
\caption{\textbf{Stability and utility must be assessed separately.}
The stability pattern spans 14 cells; utility evidence here is from separate Qwen clean pools. This does not establish a zero-cost switch or held-out predictive validity.}
\label{fig:pareto}
\end{figure}

\subsection{What is reproducible from these records}
The bundled script verifies input hashes, cell/problem uniqueness, flip/index consistency, the actual last-slot rule, comparison-safe eligibility, removal-only identities on retained rows, and equality of paired differences with displayed marginal differences. These are record-consistency checks. They cannot verify the truth of a missing semantic label, recover an unrecorded annotation, or establish that an unavailable edited step-score vector was generated correctly. Checksums establish file identity, not scientific validity.

\clearpage
\section{Complete Recorded Product--GM Results}
\label{app:cells}
\begin{table}[htbp]
\centering\small\begin{adjustbox}{max width=\linewidth}% Auto-generated; do not edit numbers by hand.
\begin{tabular}{lrrrrrr}
\toprule
Checkpoint & $n$ & Product & GM & $\Delta$ [95\% CI] & $p_{\rm exact}$ & $p_{\rm Holm}$ \\
\midrule
\multicolumn{7}{l}{\textbf{GSM8K}} \\
InternLM2-1.8B & 359 & 0.3259 & 0.0585 & \estci{$+0.2674$}{$[+0.2201, +0.3175]$} & $4.7\times10^{-25}$ & $6.6\times10^{-24}$ \\
InternLM2-7B & 397 & 0.1562 & 0.0655 & \estci{$+0.0907$}{$[+0.0579, +0.1234]$} & $4.4\times10^{-8}$ & $4.8\times10^{-7}$ \\
Math-Shepherd & 332 & 0.1024 & 0.0392 & \estci{$+0.0633$}{$[+0.0361, +0.0904]$} & $5.7\times10^{-6}$ & $4.6\times10^{-5}$ \\
Qwen-7B & 405 & 0.0395 & 0.0321 & \estci{$+0.0074$}{$[-0.0074, +0.0247]$} & 0.549 & 1.000 \\
RLHFlow-8B & 390 & 0.0436 & 0.0282 & \estci{$+0.0154$}{$[-0.0077, +0.0385]$} & 0.286 & 0.859 \\
Skywork-7B & 345 & 0.2406 & 0.0957 & \estci{$+0.1449$}{$[+0.1014, +0.1913]$} & $1.8\times10^{-10}$ & $2.2\times10^{-9}$ \\
Skywork-1.5B & 328 & 0.2195 & 0.1159 & \estci{$+0.1037$}{$[+0.0640, +0.1433]$} & $6.2\times10^{-7}$ & $6.2\times10^{-6}$ \\
\midrule
\multicolumn{7}{l}{\textbf{MATH-500}} \\
InternLM2-1.8B & 285 & 0.3123 & 0.1333 & \estci{$+0.1789$}{$[+0.1263, +0.2316]$} & $1.0\times10^{-10}$ & $1.3\times10^{-9}$ \\
InternLM2-7B & 307 & 0.2020 & 0.0977 & \estci{$+0.1042$}{$[+0.0586, +0.1498]$} & $9.1\times10^{-6}$ & $6.3\times10^{-5}$ \\
Math-Shepherd & 289 & 0.1938 & 0.1107 & \estci{$+0.0830$}{$[+0.0415, +0.1280]$} & $3.9\times10^{-4}$ & 0.0019 \\
Qwen-7B & 319 & 0.0815 & 0.0690 & \estci{$+0.0125$}{$[-0.0157, +0.0408]$} & 0.503 & 1.000 \\
RLHFlow-8B & 325 & 0.1262 & 0.0523 & \estci{$+0.0738$}{$[+0.0369, +0.1108]$} & $1.2\times10^{-4}$ & $7.0\times10^{-4}$ \\
Skywork-7B & 284 & 0.1831 & 0.0986 & \estci{$+0.0845$}{$[+0.0423, +0.1303]$} & $3.9\times10^{-4}$ & 0.0019 \\
Skywork-1.5B & 257 & 0.2101 & 0.0895 & \estci{$+0.1206$}{$[+0.0739, +0.1712]$} & $3.1\times10^{-6}$ & $2.8\times10^{-5}$ \\
\bottomrule
\end{tabular}
\end{adjustbox}
\caption{\textbf{Full recorded support.} Product and GM flip rates, paired differences, pointwise percentile intervals, exact two-sided paired $p$-values, and Holm-adjusted $p$-values. The family contains the 14 displayed cells. Rounding follows the full-precision output; individual rounded marginal differences can differ from the rounded paired estimate by one last digit.}
\label{tab:cellsfull}
\end{table}
\FloatBarrier
\begin{table}[htbp]
\centering\small\begin{adjustbox}{max width=\linewidth}% Auto-generated; do not edit numbers by hand.
\begin{tabular}{lrrrrrr}
\toprule
Checkpoint & $n$ & Product & GM & $\Delta$ [95\% CI] & $p_{\rm exact}$ & $p_{\rm Holm}$ \\
\midrule
\multicolumn{7}{l}{\textbf{GSM8K}} \\
InternLM2-1.8B & 347 & 0.3199 & 0.0432 & \estci{$+0.2767$}{$[+0.2305, +0.3228]$} & $6.2\times10^{-28}$ & $8.7\times10^{-27}$ \\
InternLM2-7B & 385 & 0.1403 & 0.0416 & \estci{$+0.0987$}{$[+0.0701, +0.1299]$} & $7.5\times10^{-11}$ & $8.2\times10^{-10}$ \\
Math-Shepherd & 328 & 0.0915 & 0.0305 & \estci{$+0.0610$}{$[+0.0335, +0.0884]$} & $1.1\times10^{-5}$ & $6.6\times10^{-5}$ \\
Qwen-7B & 397 & 0.0302 & 0.0176 & \estci{$+0.0126$}{$[+0.0025, +0.0252]$} & 0.0625 & 0.125 \\
RLHFlow-8B & 375 & 0.0267 & 0.0053 & \estci{$+0.0213$}{$[+0.0053, +0.0400]$} & 0.0215 & 0.0645 \\
Skywork-7B & 334 & 0.2275 & 0.0808 & \estci{$+0.1467$}{$[+0.1048, +0.1886]$} & $5.9\times10^{-12}$ & $7.3\times10^{-11}$ \\
Skywork-1.5B & 316 & 0.2152 & 0.0949 & \estci{$+0.1203$}{$[+0.0854, +0.1582]$} & $7.5\times10^{-11}$ & $8.2\times10^{-10}$ \\
\midrule
\multicolumn{7}{l}{\textbf{MATH-500}} \\
InternLM2-1.8B & 273 & 0.3040 & 0.1172 & \estci{$+0.1868$}{$[+0.1355, +0.2381]$} & $5.6\times10^{-12}$ & $7.3\times10^{-11}$ \\
InternLM2-7B & 294 & 0.1769 & 0.0782 & \estci{$+0.0986$}{$[+0.0578, +0.1395]$} & $9.0\times10^{-6}$ & $6.3\times10^{-5}$ \\
Math-Shepherd & 274 & 0.1679 & 0.0985 & \estci{$+0.0693$}{$[+0.0328, +0.1095]$} & $8.8\times10^{-4}$ & 0.0035 \\
Qwen-7B & 302 & 0.0563 & 0.0364 & \estci{$+0.0199$}{$[+0.0033, +0.0397]$} & 0.0703 & 0.125 \\
RLHFlow-8B & 314 & 0.1115 & 0.0287 & \estci{$+0.0828$}{$[+0.0510, +0.1178]$} & $8.7\times10^{-7}$ & $7.8\times10^{-6}$ \\
Skywork-7B & 273 & 0.1612 & 0.0842 & \estci{$+0.0769$}{$[+0.0366, +0.1209]$} & $5.1\times10^{-4}$ & 0.0025 \\
Skywork-1.5B & 247 & 0.1903 & 0.0810 & \estci{$+0.1093$}{$[+0.0648, +0.1538]$} & $3.5\times10^{-6}$ & $2.8\times10^{-5}$ \\
\bottomrule
\end{tabular}
\end{adjustbox}
\caption{\textbf{Comparison-safe support.} The same fields on pre-injection-defined retained rows. Every positive percentile interval should be read alongside the exact tests, particularly Qwen. The family-wise adjustment is across these 14 cells, not all analyses in the paper.}
\label{tab:cellssafe}
\end{table}
\FloatBarrier

\subsection{Discordance counts and secondary reducers}
The next tables expose the effective paired-test sample. Each discordance entry is product-only/comparator-only. They are not the total flip counts: both reducers can flip on the same question and contribute zero to the paired difference.
\begin{table}[htbp]
\centering\small\begin{adjustbox}{max width=\linewidth}% Auto-generated; do not edit numbers by hand.
\begin{tabular}{lrrrrr}
\toprule
Checkpoint & Full discordance & Exact $p$ & Safe discordance & Exact $p$ & Safe Holm $p$ \\
\midrule
\multicolumn{6}{l}{\textbf{GSM8K}} \\
InternLM2-1.8B & 100/4 & $4.7\times10^{-25}$ & 97/1 & $6.2\times10^{-28}$ & $8.7\times10^{-27}$ \\
InternLM2-7B & 41/5 & $4.4\times10^{-8}$ & 39/1 & $7.5\times10^{-11}$ & $8.2\times10^{-10}$ \\
Math-Shepherd & 22/1 & $5.7\times10^{-6}$ & 21/1 & $1.1\times10^{-5}$ & $6.6\times10^{-5}$ \\
Qwen-7B & 7/4 & 0.549 & 5/0 & 0.0625 & 0.125 \\
RLHFlow-8B & 14/8 & 0.286 & 9/1 & 0.0215 & 0.0645 \\
Skywork-7B & 58/8 & $1.8\times10^{-10}$ & 53/4 & $5.9\times10^{-12}$ & $7.3\times10^{-11}$ \\
Skywork-1.5B & 41/7 & $6.2\times10^{-7}$ & 39/1 & $7.5\times10^{-11}$ & $8.2\times10^{-10}$ \\
\midrule
\multicolumn{6}{l}{\textbf{MATH-500}} \\
InternLM2-1.8B & 59/8 & $1.0\times10^{-10}$ & 56/5 & $5.6\times10^{-12}$ & $7.3\times10^{-11}$ \\
InternLM2-7B & 42/10 & $9.1\times10^{-6}$ & 36/7 & $9.0\times10^{-6}$ & $6.3\times10^{-5}$ \\
Math-Shepherd & 34/10 & $3.9\times10^{-4}$ & 25/6 & $8.8\times10^{-4}$ & 0.0035 \\
Qwen-7B & 12/8 & 0.503 & 7/1 & 0.0703 & 0.125 \\
RLHFlow-8B & 31/7 & $1.2\times10^{-4}$ & 28/2 & $8.7\times10^{-7}$ & $7.8\times10^{-6}$ \\
Skywork-7B & 34/10 & $3.9\times10^{-4}$ & 28/7 & $5.1\times10^{-4}$ & 0.0025 \\
Skywork-1.5B & 38/7 & $3.1\times10^{-6}$ & 31/4 & $3.5\times10^{-6}$ & $2.8\times10^{-5}$ \\
\bottomrule
\end{tabular}
\end{adjustbox}
\caption{Product versus GM discordance counts on full and comparison-safe support.}
\end{table}
\FloatBarrier
\begin{table}[htbp]
\centering\small\begin{adjustbox}{max width=\linewidth}% Auto-generated; do not edit numbers by hand.
\begin{tabular}{lrrrrr}
\toprule
Checkpoint & Full discordance & Exact $p$ & Safe discordance & Exact $p$ & Safe Holm $p$ \\
\midrule
\multicolumn{6}{l}{\textbf{GSM8K}} \\
InternLM2-1.8B & 90/14 & $9.2\times10^{-15}$ & 87/10 & $1.8\times10^{-16}$ & $2.3\times10^{-15}$ \\
InternLM2-7B & 43/7 & $2.1\times10^{-7}$ & 37/5 & $4.4\times10^{-7}$ & $4.0\times10^{-6}$ \\
Math-Shepherd & 21/2 & $6.6\times10^{-5}$ & 17/2 & $7.3\times10^{-4}$ & 0.0051 \\
Qwen-7B & 10/0 & 0.0020 & 7/0 & 0.0156 & 0.0677 \\
RLHFlow-8B & 13/1 & 0.0018 & 7/1 & 0.0703 & 0.196 \\
Skywork-7B & 61/6 & $1.5\times10^{-12}$ & 54/6 & $9.7\times10^{-11}$ & $1.2\times10^{-9}$ \\
Skywork-1.5B & 45/4 & $8.2\times10^{-10}$ & 42/3 & $8.7\times10^{-10}$ & $9.5\times10^{-9}$ \\
\midrule
\multicolumn{6}{l}{\textbf{MATH-500}} \\
InternLM2-1.8B & 68/8 & $5.6\times10^{-13}$ & 64/2 & $6.0\times10^{-17}$ & $8.4\times10^{-16}$ \\
InternLM2-7B & 46/8 & $1.4\times10^{-7}$ & 40/6 & $3.1\times10^{-7}$ & $3.1\times10^{-6}$ \\
Math-Shepherd & 37/8 & $1.5\times10^{-5}$ & 27/8 & 0.0019 & 0.0113 \\
Qwen-7B & 18/4 & 0.0043 & 10/4 & 0.180 & 0.196 \\
RLHFlow-8B & 32/4 & $1.9\times10^{-6}$ & 27/4 & $3.4\times10^{-5}$ & $2.7\times10^{-4}$ \\
Skywork-7B & 32/9 & $4.3\times10^{-4}$ & 24/9 & 0.0135 & 0.0677 \\
Skywork-1.5B & 30/12 & 0.0079 & 24/12 & 0.0652 & 0.196 \\
\bottomrule
\end{tabular}
\end{adjustbox}
\caption{Secondary product versus min-step discordances. The full-support and retained-support inference can differ materially. In particular, the retained MATH-500 Qwen comparison is inconclusive rather than a universal product-to-min recommendation.}
\end{table}
\FloatBarrier
\begin{table}[htbp]
\centering\small\begin{adjustbox}{max width=\linewidth}% Auto-generated; do not edit numbers by hand.
\begin{tabular}{lrrrrr}
\toprule
Checkpoint & Full discordance & Exact $p$ & Safe discordance & Exact $p$ & Safe Holm $p$ \\
\midrule
\multicolumn{6}{l}{\textbf{GSM8K}} \\
InternLM2-1.8B & 112/5 & $2.1\times10^{-27}$ & 107/5 & $5.4\times10^{-26}$ & $7.6\times10^{-25}$ \\
InternLM2-7B & 61/6 & $1.5\times10^{-12}$ & 53/6 & $1.8\times10^{-10}$ & $1.9\times10^{-9}$ \\
Math-Shepherd & 27/9 & 0.0039 & 25/3 & $2.7\times10^{-5}$ & $1.4\times10^{-4}$ \\
Qwen-7B & 16/2 & 0.0013 & 13/0 & $2.4\times10^{-4}$ & $9.8\times10^{-4}$ \\
RLHFlow-8B & 16/3 & 0.0044 & 10/1 & 0.0117 & 0.0117 \\
Skywork-7B & 80/5 & $1.8\times10^{-18}$ & 73/0 & $2.1\times10^{-22}$ & $2.8\times10^{-21}$ \\
Skywork-1.5B & 67/7 & $2.1\times10^{-13}$ & 64/1 & $3.6\times10^{-18}$ & $4.3\times10^{-17}$ \\
\midrule
\multicolumn{6}{l}{\textbf{MATH-500}} \\
InternLM2-1.8B & 66/32 & $7.7\times10^{-4}$ & 62/32 & 0.0026 & 0.0067 \\
InternLM2-7B & 50/17 & $6.7\times10^{-5}$ & 41/17 & 0.0022 & 0.0067 \\
Math-Shepherd & 46/10 & $1.2\times10^{-6}$ & 36/6 & $2.8\times10^{-6}$ & $2.0\times10^{-5}$ \\
Qwen-7B & 26/0 & $3.0\times10^{-8}$ & 18/0 & $7.6\times10^{-6}$ & $4.6\times10^{-5}$ \\
RLHFlow-8B & 37/7 & $5.3\times10^{-6}$ & 34/3 & $1.2\times10^{-7}$ & $9.9\times10^{-7}$ \\
Skywork-7B & 47/10 & $7.5\times10^{-7}$ & 40/2 & $4.1\times10^{-10}$ & $4.1\times10^{-9}$ \\
Skywork-1.5B & 50/8 & $1.6\times10^{-8}$ & 44/4 & $1.5\times10^{-9}$ & $1.4\times10^{-8}$ \\
\bottomrule
\end{tabular}
\end{adjustbox}
\caption{Secondary product versus last-step discordances. A lower targeted flip rate does not establish better natural-pool accuracy or process quality for last-step aggregation.}
\end{table}
\FloatBarrier

\section{Necessary/Inert Recount and Question-Cluster Inference}
\label{app:ni}
\begin{table}[htbp]
\centering\small\begin{adjustbox}{max width=\linewidth}% Auto-generated; do not edit numbers by hand.
\begin{tabular}{lrrrr}
\toprule
Checkpoint & Pairs & Raw $\Delta_{\rm NI}$ [95\% CI] & Rematched pairs & Matched $\Delta_{\rm NI}$ [95\% CI] \\
\midrule
\multicolumn{5}{l}{\textbf{GSM8K}} \\
InternLM2-1.8B & 200 & \estci{$-0.1600$}{$[-0.2585, -0.0624]$} & 177 & \estci{$-0.1751$}{$[-0.2853, -0.0739]$} \\
InternLM2-7B & 200 & \estci{$-0.2550$}{$[-0.3798, -0.1336]$} & 190 & \estci{$-0.2632$}{$[-0.3907, -0.1399]$} \\
Math-Shepherd & 200 & \estci{$-0.4100$}{$[-0.5281, -0.2785]$} & 194 & \estci{$-0.4175$}{$[-0.5364, -0.2906]$} \\
Qwen-7B & 200 & \estci{$-0.3350$}{$[-0.4577, -0.2032]$} & 199 & \estci{$-0.3317$}{$[-0.4548, -0.2000]$} \\
RLHFlow-8B & 200 & \estci{$-0.3100$}{$[-0.4327, -0.1784]$} & 198 & \estci{$-0.3030$}{$[-0.4272, -0.1686]$} \\
Skywork-7B & 200 & \estci{$-0.3650$}{$[-0.5021, -0.2133]$} & 180 & \estci{$-0.3278$}{$[-0.4635, -0.1736]$} \\
Skywork-1.5B & 200 & \estci{$-0.3950$}{$[-0.5366, -0.2552]$} & 187 & \estci{$-0.4118$}{$[-0.5472, -0.2735]$} \\
\midrule
\multicolumn{5}{l}{\textbf{MATH-500}} \\
InternLM2-1.8B & 200 & \estci{$-0.0900$}{$[-0.1647, -0.0161]$} & 145 & \estci{$-0.0552$}{$[-0.1333, +0.0251]$} \\
InternLM2-7B & 200 & \estci{$-0.3400$}{$[-0.4666, -0.2076]$} & 180 & \estci{$-0.3611$}{$[-0.4892, -0.2255]$} \\
Math-Shepherd & 200 & \estci{$-0.3250$}{$[-0.4873, -0.1524]$} & 198 & \estci{$-0.3283$}{$[-0.4916, -0.1528]$} \\
Qwen-7B & 200 & \estci{$-0.1450$}{$[-0.2434, -0.0490]$} & 199 & \estci{$-0.1457$}{$[-0.2449, -0.0499]$} \\
RLHFlow-8B & 200 & \estci{$-0.3050$}{$[-0.4466, -0.1518]$} & 183 & \estci{$-0.2568$}{$[-0.3773, -0.1201]$} \\
Skywork-7B & 200 & \estci{$-0.2900$}{$[-0.4222, -0.1491]$} & 195 & \estci{$-0.2923$}{$[-0.4214, -0.1527]$} \\
Skywork-1.5B & 200 & \estci{$-0.3400$}{$[-0.4593, -0.2156]$} & 190 & \estci{$-0.3632$}{$[-0.4869, -0.2383]$} \\
\bottomrule
\end{tabular}
\end{adjustbox}
\caption{\textbf{NI estimates with question-cluster intervals.} Point estimates and counts are recomputed from the public per-pair score records and equal the pinned \texttt{answer\_preserved\_ni.csv} values exactly. Intervals resample question IDs (10,000 multinomial replicates, PCG64 seed 42), carrying all necessary and inert instances of a sampled question together and holding fixed which instances enter each arm (the pairs and, for the matched column, the historical caliper assignment). They replace the historical pair-row intervals.}
\label{tab:ni}
\end{table}

The historical script \filepath{code/original_scripts/920_answer_preserved_ni.py} passes a vector of paired differences to a bootstrap that samples its row indices; it does not pass the source question IDs into that function. The public table printer \filepath{reproduce_table1.py} only reads and prints the result CSV. This version instead recomputes the statistics from the pair-level records of the pinned repository. The files \filepath{scores/*_necessary.jsonl} and \filepath{scores/*_inert.jsonl} store, for every instance, the question ID, candidate index, deleted step, and the clean and edited step-score vectors; \filepath{edits/*_matched.jsonl} store the texts used for the answer-preservation filter, which retains all 200 pairs per benchmark. \filepath{anc/derive_ni_records.py} checks these 32 files against their Git blob identities and extracts \filepath{anc/inputs/ni_pair_records.csv}. All 14 raw and score-matched estimates and matched counts are reproduced exactly, and the product indicator recomputed from the step-score vectors agrees with the stored trajectory scores on all 5,600 checkpoint-level records (seven checkpoints $\times$ 800 deletion instances).

All 14 raw and score-matched point estimates are negative. With question-cluster resampling, the 95\% interval lies below zero in all 14 raw cells and, under the historical assignment, in 13 of 14 score-matched cells; the exception is InternLM2-1.8B on MATH-500, $-0.055$ $[-0.133,+0.025]$. These are the counts the earlier version reported from pair-row intervals, so the conclusion survives the corrected resampling unit, but the intervals are 1.3--1.9 times as wide (median 1.5). The 70/70 five-seed sign pattern and the five-seed ranges are not retained: the public multi-seed score files pool 815 (GSM8K) or 897 (MATH-500) distinct pairs per cell without seed labels, so the per-seed samples cannot be reconstructed, and the listed seed-42 value is not the primary 200-pair estimate.

The intervals hold fixed which instances enter each arm: the constructed pairs, the answer-preservation filter, and, for the matched estimate, the caliper selection. Because the two members of a pair come from different questions in 399 of the 400 pairs, the bootstrap reweights each arm through its questions; the pairing itself does not enter, since $\Delta_{\mathrm{NI}}$ is a difference of the two arm means. The intervals do not propagate uncertainty from redrawing the matches or relabeling necessity.

The caliper selection is, moreover, not unique. Its objective adds the distances of the pairs left outside the caliper, and for one-dimensional distances many reassignments leave that total unchanged, so many assignments attain exactly the optimal cost. All of them have the same number of matched pairs, because the penalty forces the maximum number of within-caliper pairs. The historical assignment is reproduced with its original floating-point arithmetic and row order. For exact sensitivity, \filepath{anc/reproduce.py} fixes the historical caliper mask and treats stored binary64 log-bonuses as exact dyadic rationals before forming penalized distances. It verifies that the historical assignment is optimal for this objective and bounds the matched estimate over all its optima; this does not enumerate ties of a separately rounded cost matrix. The estimate is unique in four cells (Qwen-7B on both benchmarks, RLHFlow-8B on GSM8K, and Math-Shepherd on MATH-500) and negative at every optimum in 13 cells; for InternLM2-1.8B on MATH-500 it ranges from $-0.138$ to $+0.028$. For one selected point-estimate-maximizing assignment per cell, 12/14 pointwise intervals are below zero; InternLM2-1.8B/GSM8K gives $[-0.1896,+0.0052]$. Assignments with equal point estimates can have different question composition and intervals. Interval endpoints were not optimized over all optima; 12/14 is therefore not a worst-case or simultaneous significance guarantee. Table~\ref{tab:ni} reports the historical assignment; \filepath{anc/outputs/ni_cluster.csv} contains both exact point-estimate extremes and intervals for selected representative assignments. The raw estimates do not use the assignment and are unaffected. A negative point estimate is not a substitute for calibrated uncertainty, and these intervals do not identify a causal effect of aggregation.

\FloatBarrier
\section{Human Endpoints, Ethics, and Independence}
\label{app:human}
\subsection{Three-choice reannotation}
\begin{table}[htbp]
\centering\small% Auto-generated; do not edit numbers by hand.
\begin{tabular}{lrrrrr}
\toprule
Subset & Cards & GM & No preference & Product & Split \\
\midrule
InternLM2 / GSM8K & 50 & 37 & 12 & 1 & 0 \\
InternLM2 / MATH & 50 & 25 & 25 & 0 & 0 \\
Skywork / GSM8K & 50 & 30 & 20 & 0 & 0 \\
Skywork / MATH & 50 & 21 & 27 & 0 & 2 \\
\midrule
All cards & 200 & 113 & 84 & 1 & 2 \\
\bottomrule
\end{tabular}

\caption{\textbf{All outcomes on 200 selected reannotation cards.} Four non-overlapping strata sum to the total. ``Split'' means no category obtains a majority. Counts are from a pinned published summary, not a new annotator-level replication.}
\end{table}

The cards compare answer-correct traces selected differently by product and GM under deletion stress. A/B presentation was blinded and randomized according to the study protocol; the summary is based on majority outcomes with a no-preference option. The 200-card subset was reannotated after an earlier forced-choice study. Calling it the leading \emph{descriptive} endpoint here does not make it an originally preregistered primary endpoint.

The retained distribution is 56.5\% GM, 42.0\% no preference, 0.5\% product, and 1.0\% split. A resolved-card rate would exclude neutral and split cards and answer a different question. We do not use a high resolved-card percentage or the earlier 85.5\% forced-choice headline to represent all cards, all original pools, or all PRMs. Length, task selection, and prior annotator exposure may influence preference. No length-independent causal interpretation is supported.

\subsection{Natural-pool reannotation}
\begin{table}[htbp]
\centering\small\begin{adjustbox}{max width=\linewidth}% Auto-generated; do not edit numbers by hand.
\begin{tabular}{lrrrrr}
\toprule
Subset & Cards & Comparator & Product & No preference & Split \\
\midrule
Skywork-7B / GSM8K & 70 & 1 & 1 & 68 & 0 \\
Skywork-7B / MATH-500 & 70 & 6 & 2 & 61 & 1 \\
InternLM2-7B / GSM8K & 70 & 6 & 1 & 63 & 0 \\
InternLM2-7B / MATH-500 & 70 & 15 & 8 & 45 & 2 \\
Qwen-7B / GSM8K & 70 & 0 & 0 & 70 & 0 \\
Qwen-7B / MATH-500 & 70 & 0 & 3 & 67 & 0 \\
\midrule
All cards & 420 & 28 & 15 & 374 & 3 \\
\bottomrule
\end{tabular}
\end{adjustbox}
\caption{\textbf{Natural-pool human summaries.} ``Comparator'' is the comparator label stored by the source study; it is not reinterpreted as an endorsement of a model release's official reducer. These are different cards and a different comparison regime from the deletion-stress reannotation.}
\end{table}

There are 374 no-preference majorities, or 89.05\%, and 43 resolved preference majorities. Among the latter, 28 favor the stored comparator and 15 product; the two-sided exact binomial test is approximately $p=0.066$. This is descriptive evidence with a substantial neutral majority, not proof of a deployment-wide process-quality benefit. Evaluators may overlap across substudies, so these endpoints are not independent replications of one another.

\subsection{Expert mini-replication: excluded from independent evidence}
The frozen repository's \filepath{human_cards/expert_replication/input_audit.txt}, dated 15 April 2026, states that the three completed expert workbooks are byte-identical, with identical choices on all 40 cards. It explicitly limits the available evidence to at most one distinct expert return. The historical summary and later manuscript counts also disagree. This version does not choose the more favorable set of counts or infer three independent experts from three filenames.

Accordingly, this version does not claim that three distinct workbooks were verified or that an independent three-expert mini-replication corroborates the result. Conflicting expert-rate summaries are excluded from empirical support. The original audit is preserved in \filepath{anc/provenance/expert_input_audit.txt}. Its existence does not prove that no independent study ever occurred; it means the available materials do not establish one. A future correction requires genuinely distinct original returns and a dated provenance mapping, not removal of the audit history.

\subsection{Other human results}
Qwen Likert multi-rater estimates, some Likert differences, and expert agreement measures are not relied upon where their completed-workbook archive cannot be verified. This does not invalidate the two descriptive summary tables above by logical implication, but it prevents treating the broader human package as uniformly independently validated. The reanalysis here verifies summary arithmetic and file identity only.

\subsection{Ethics and recruitment}
The study did not undergo formal review by an institutional review board or ethics committee. No claim of an institutional exemption determination is made. The study protocol reports adult participation, informed consent, compensation for quantitative annotation work, de-identified mathematical text, and exclusion of personal or sensitive information from the cards. Uncompensated pilot feedback was used for rubric development rather than quantitative endpoints. The original study description reports a separate edit-labeling pool and a shared quality-evaluation pool, with some participation across substudies; complete source-specific recruitment and temporal records are not recoverable from the available archive.

Only de-identified outcome summaries are included in this paper's human-analysis inputs. No new human annotation was performed for this version. A license or public repository location does not establish participant consent for releasing personally identifying metadata; future releases must separately verify the consent and de-identification of original workbooks.

\section{Evidence Availability and Claim Disposition}
\label{app:status}
This version distinguishes unavailable evidence from a proven false result. A numerical claim is not made more reliable merely by carrying it into a new PDF or a hash manifest. The following ledger specifies how each substantive issue is handled.

\paragraph{Retained and recomputed from outcome records.}
The full 14-cell product--GM point estimates, recorded product--min/last comparisons, comparison-safe subsets, Qwen $N=32$ natural utility, and MATH difficulty statistics are regenerated from pinned per-problem selector records. The 72B pilot is recomputed from its 143-row public CSV. The raw and score-matched NI point estimates for all 14 cells are recomputed from the public per-pair score records, and their intervals are replaced by question-cluster intervals; the score-matching assignment is shown to be one of many exact optima, and the matched estimate is bounded over all of them (Appendix~\ref{app:ni}). These are paired-outcome analyses, not renewed PRM inference. The original 14-cell macro point remains $0.0964616076513$; rounding it directly gives 0.096 at three decimals or 0.0965 at four decimals.

\paragraph{Figure and table corrections.}
Figure~\ref{fig:forest} is regenerated from the same result table as the numerical appendix. Its InternLM2-1.8B/GSM8K point is 0.2674095, not the inconsistent earlier plotted value. Corrected MATH marginal rates use the same common support as the paired delta. The named six-cell Qwen/Skywork subset is not described as the five-native-PRM population. Additional earlier prevalence, NI amplification, and landscape tables whose complete derivations are not packaged are not used to support new claims in this version.

\paragraph{Preserved as descriptive summaries.}
The two human majority-count tables are retained with the limitations described in Appendix~\ref{app:human}. Their source summary files are included, with provenance. They are not represented as independently recovered annotation returns. Historical summary fields that remain inside an archived CSV, such as the pair-row NI intervals, are not automatically asserted by this paper.

\paragraph{Replacement wording.}
The actual last non-default-winner slot replaces the incorrect lowest-score description. Comparison-safe filtering is a conditional sensitivity analysis, not a correction that retroactively changes historical outcomes. Alternative injection protocols remain outside the verified evidence.

\paragraph{Withheld $N=8$ and multi-seed claims.}
The available $N=8$ summary does not include its per-question subsampling manifest or 20-seed output archive. We do not infer seed-wise counts or decision stability from a summary status label. The earlier five-seed NI range also conflicts with the listed seed-42 value, and the public multi-seed score files do not record which pairs belong to which seed. Neither claim is retained as robustness evidence. Recovery requires the actual input identities, scripts/configuration, and all seed-level outputs, including unfavorable results.

\paragraph{72B scope correction.}
Only the 200-question GSM8K sample with 143 eligible records is available in the pinned 72B directory. The former 600-question two-benchmark pooled claim and its significant CI are withdrawn. The retained pilot's exact $p=0.125$ and zero-touching interval do not support a scale-general conclusion.

\paragraph{Missing auxiliary empirical archives.}
Claims about quantified non-strict inflation by $0.01\times$--$11\times$, DeepSeek replication, a ReasonFlux mixed ordering, new trimmed/median/length-cost reducer performance, symmetric attack-source 18/18 results, $N$-sweeps, selected placeholder results, and some step-level, optimization, and Likert tables are not counted as validated evidence in this version without their corresponding inputs and outputs. The underlying methodological questions remain meaningful; their former numerical answers are not replaced by invented ones. Generic reducer definitions and the mechanical derivation remain available as methods, not experimental wins.

\paragraph{Reproducibility claims narrowed.}
The original repository contains both cached vectors and precomputed summaries. Some reproduction entrypoints print existing tables rather than rebuilding them. The primary edited-score vectors needed to evaluate a new reducer on the identical stress pool are not available in the inspected materials. This version therefore claims complete CPU reanalysis of its \emph{included outcome records}, not unrestricted exact replay of every old table or every new reducer.

\paragraph{Licensing and release status.}
The pinned root has a short CC-BY-4.0 marker. Dedicated data/code license texts, a NOTICE, and third-party manifests previously described as present are not asserted to exist in that old snapshot. No blanket MIT code grant or relicensing of upstream models/datasets is inferred. A future repository release may add valid rights documentation and missing evidence, but that must be a new, traceable version. The revision does not alter the public repository or erase historical audits.

\section{Executable Artifact Contract}
\label{app:artifact}
\subsection{Versioned inputs}
The original repository commit is
\begin{quote}\small\path{2ad65edd36e93c5141f5d091f0c23b9e38309395}.\end{quote}
The corresponding dataset snapshot referenced by the supplied analysis records is
\begin{quote}\small\path{555e90adc36db47019797510bf7c6fa64565d181}.\end{quote}
The accompanying \filepath{anc/input_manifest.json} records exact SHA-256 identities and the origin of every input. Public CSV snapshots additionally carry their Git blob identity where verified. Selector exports from the supplied analysis bundle are identified as such, not falsely relabeled as raw model outputs from a new run. Absolute machine paths and obsolete internal decision labels are not needed to reproduce the numerical calculations.

\begin{table}[htbp]
\centering\small
\begin{tabularx}{\linewidth}{L{5.0cm}Y}
\toprule
Input in \texttt{anc/inputs/} & Evidence role\\
\midrule
\filepath{selection_records.csv} & Recorded clean/injected identities and paired flip outcomes; all three comparators.\\
\filepath{qwen_natural_records.csv} & Recorded Qwen clean selections and correctness for 500 pools per benchmark.\\
\filepath{math_level_records.csv} & Level-stratification records, clean-correct flags, and original-candidate checks.\\
\filepath{math500_level_map.csv} & One-to-one problem/level mapping and source-question hashes.\\
\filepath{qwen72b_records.csv} & Public 143-row GSM8K pilot.\\
\filepath{historical_ni_summary.csv} & Published 14-cell summary; recount target for the NI estimates.\\
\filepath{ni_pair_records.csv} & One row per checkpoint and NI deletion instance ($7\times800=5{,}600$), extracted from 32 public per-pair score and edit files: question and candidate IDs, product indicator, deleted-step clean score.\\
\filepath{human_three_choice_summary.csv} & Published 200-card majority summary.\\
\filepath{human_natural_summary.csv} & Published 420-card majority summary.\\
\bottomrule
\end{tabularx}
\caption{Included evidence has different verification levels. Outcome fields are not a substitute for score vectors, labels, or raw annotator returns.}
\end{table}

\subsection{Commands and outputs}
From the extracted source directory, the numerical reanalysis is:
\begin{quote}\small\ttfamily
python -m pip install -r anc/requirements.txt\\
python anc/reproduce.py\\
python anc/make\_tables.py
\end{quote}
The script verifies the ten pinned inputs (nine data files plus the expert audit), reconstructs the paired statistics, and writes cell, pooled, utility, level, and NI outputs. \filepath{anc/derive_ni_records.py} optionally regenerates the NI input from a checkout of the pinned repository. Generated LaTeX tables are stored in \texttt{tables/}. The forest figure is produced from the same cell-results file by \filepath{anc/make_figure.py}. The manuscript is built with \texttt{pdflatex}, without shell escape; the Python analyses are not executed by the TeX compiler.

The numerical script requires no model weights, GPU, network access, account credentials, or access to the original author's machine. Environment versions are recorded in \filepath{anc/outputs/environment.json}. Hash verification fails on altered inputs rather than silently accepting a different release. Output rounding is applied only after estimation. A successful script run is a verification of its stated record contract, not a certification of every historical study procedure.

\subsection{How to evaluate a genuinely new reducer}
A new deterministic scalar reducer can be applied without model inference \emph{if all required clean and edited score vectors are present}. The correct procedure is to keep pool membership, candidate identity, edit identity, score extraction, and tie-breaking fixed; apply the new function to every candidate vector; recompute clean and injected winners; then evaluate on a clearly defined common support. Simply substituting a formula into a CSV of existing winners cannot recover the missing new winners. The NI control already meets the condition: both vectors are public for all 800 NI deletion instances under each of the seven checkpoints, so a new reducer's $\Delta_{\mathrm{NI}}$ can be computed exactly. The selection-stress replay is what remains blocked.

For missing edited score vectors, one must recover the original vectors or rescore the exact edited texts with the declared checkpoint and input preprocessing. If the model, tokenizer, generation, or edit changes, the result belongs to a new documented run, not exact replay of the historical evidence. A learned or trajectory-aware selector that consumes additional information likewise requires an extension protocol and its own training/inference provenance. No output of such a selector is fabricated here.

\subsection{Release maintenance}
Corrections should be additive and versioned. Preserve the frozen source and an audit log; identify which tables are superseded and why; publish new hashes for new evidence rather than pretending it was present at the old commit. Original audit records are retained even when newer files later repair an archive error. No unverified timeline is described as formal preregistration.

\section{Model Provenance and Evaluation Scope}
\label{app:models}
\begin{table}[htbp]
\centering\small
\begin{tabularx}{\linewidth}{L{3.2cm}L{3cm}Y}
\toprule
Checkpoint label & Base family/scale & Role in this benchmark\\
\midrule
Math-Shepherd & Mistral, 7B & Process scorer; stepwise verification~\citep{wang2024mathshepherd}.\\
Skywork-7B & Qwen, 7B & Process scorer~\citep{skyworkprm2024}.\\
Skywork-1.5B & Qwen, 1.5B & Process scorer~\citep{skyworkprm2024}.\\
Qwen-7B & Qwen, 7B & Qwen2.5-Math PRM; documented product aggregation~\citep{qwen25mathprm2025,qwenprmblog2024}.\\
RLHFlow-8B & Llama, 8B & MC-estimation process reward implementation, not a 7B RLHF checkpoint~\citep{rlhflow2024}.\\
InternLM2-7B & InternLM2, 7B & Scalar reward checkpoint adapted by prefix scoring~\citep{internlm2reward2024}.\\
InternLM2-1.8B & InternLM2, 1.8B & Scalar reward checkpoint adapted by prefix scoring~\citep{internlm2reward2024}.\\
\bottomrule
\end{tabularx}
\caption{Checkpoint labels are identities, not a claim of homogeneous training or identical release-default reducers. Upstream model licenses remain separate from the manuscript and analysis-record rights.}
\end{table}

The benchmark uses a common harness-default source trace to define the omission attack. In the original harness, the two InternLM2 prefix adaptations use last-step selection and the five process scorers use min-step for that source selection. This harness choice is not necessarily the reducer recommended by each model release. In particular, Qwen's documented product choice is not relabeled as official min-step. The selected attack source need not coincide with the product or GM winner.

The fixed six-cell subset previously labeled \texttt{native\_only} comprises Qwen-7B, Skywork-7B, and Skywork-1.5B across both datasets. It does not include every native PRM. Its full-support mean is $+0.0789437$, and its question-bootstrap CI conditional on those six cells is approximately $[0.0629,0.0955]$. This named subset cannot establish a training-provenance interaction or the effect of removing all adapted models from a five-native-PRM population.

The Qwen-72B pilot uses the same recorded-selection analysis concept but is limited to its available GSM8K sample. Neither it nor the Qwen-7B boundary case establishes that product aggregation was part of the checkpoint's training objective. Training provenance and inference-time reduction must be kept separate.

\section{Datasheet, Intended Use, and Extension Checklist}
\label{app:datasheet}
\paragraph{Motivation and unit.}
The resource supports auditing how a trajectory reducer changes selection on fixed scored mathematical reasoning candidates. The principal unit for paired uncertainty is a problem within a checkpoint; the pooled analysis carries shared problem observations across checkpoints. A human card, an edited trace pair, a candidate pool, and a checkpoint--problem row are distinct objects.

\paragraph{Composition.}
The original benchmark describes frozen 32-trace pools for GSM8K/MATH-500, clean scores and edit metadata for seven checkpoints, necessary/inert matching records, and human cards/summaries. The reanalysis accompanying this version includes the narrower, explicitly enumerated records in Appendix~\ref{app:artifact}. It does not include an archive that was never supplied merely because the manuscript once described it.

\paragraph{Generation and selection.}
The original study reports QwQ-32B-Preview generation with temperature 0.7, top-$p$ 0.9, $N=32$, and a 4,096-token limit. These are historical construction settings, not independently rerun in the present CPU analysis. The stress sample is selected by edit and scorer eligibility; it is not a random deployment sample. Restricting to answer-preserving or unambiguous omissions can affect the population to which conclusions apply.

\paragraph{Intended uses.}
Appropriate uses include reproducing recorded reducer comparisons, diagnosing replacement confounding, checking sensitivity to outcome support, and designing transparent future reducer experiments. Inappropriate uses include ranking all PRMs by deployment safety, treating a lower Flip rate as sufficient proof of better reasoning, or asserting that the reported robustness score predicts held-out accuracy.

\paragraph{Missingness.}
Record absence is not a zero outcome. Missing primary edited-score vectors limit new-reducer stress replay. Missing raw annotations limit independence and agreement claims. Missing seed-level logs limit robustness claims. These categories require different remedies and cannot be repaired by the same summary statistic or checksum.

\paragraph{Privacy and rights.}
The included analytical records contain problem identifiers, model labels, selections, correctness flags, and aggregate human outcomes, not annotator identities. The original model, problem-dataset, code, and annotation rights remain distinct. A manuscript license is not a blanket license for all upstream material. This version makes no representation that absent third-party notices or dedicated code/data license documents exist in the pinned repository.

\paragraph{Cross-domain extension.}
The reusable elements are frozen pools, a declared process unit, matched interventions, identical scored evidence across reducers, explicit tie-breaking, paired uncertainty, and versioned provenance. The domain-specific elements are the definition of a necessary omission, the preservation oracle, the local-error verifier, and task/environment correctness. Code requires execution and tests with AST-aware edits~\citep{evalplus2024}; proofs require a proof checker and local obligations; agents require replayable tool logs, tool outputs, and environment state. None is claimed as an empirically validated extension here.

\paragraph{Minimum evidence for a future extension.}
Store both the original and edited process, the verification result, every candidate's required scores, source/checkpoint/tokenizer versions, the attacked and replaced identities, before/after selections, and task correctness. Record exclusions and all seeds, not only favorable runs. Use separate held-out tasks for any claim that tuning a reducer on omission stability improves task utility. A trained selector must not be evaluated as though it were a fixed, untuned scalar function.

\paragraph{Maintenance and correction policy.}
The goal is to make supported conclusions inspectable and unsupported conclusions visibly absent. Corrections to manuscript figures or tables should use generated outputs from a pinned source, with a dated change record. Claims may be restored when the required genuine evidence is recovered and verified; their absence from this version is not a promise that such records can be reconstructed. The original expert audit and the versioned source history are retained rather than overwritten.

\end{document}